\documentclass[nohyperref]{article}

\usepackage[utf8]{inputenc} 
\usepackage[T1]{fontenc}    
\usepackage{url}            
\usepackage{booktabs}       
\usepackage{amsfonts}       
\usepackage{nicefrac}       
\usepackage{microtype}      
\usepackage{soul}
\usepackage{graphicx}
\usepackage{amsmath}
\usepackage{outlines}
\usepackage{xurl}

\usepackage[colorlinks=true,
citecolor=blue,
filecolor=black,
linkcolor=blue,
urlcolor=blue]{hyperref}

\usepackage{amsmath,amsfonts,bm}

\usepackage{xcolor}
\usepackage{colortbl}

\def\eqref#1{equation~\ref{#1}}

\DeclareMathAlphabet{\mathsfit}{\encodingdefault}{\sfdefault}{m}{sl}
\SetMathAlphabet{\mathsfit}{bold}{\encodingdefault}{\sfdefault}{bx}{n}

\def\sR{{\mathbb{R}}}

\newcommand{\sigmoid}{\sigma}

\usepackage{multirow}
\usepackage{paralist}

\usepackage{multicol}
\usepackage{diagbox}
\usepackage{arydshln}
\usepackage[ruled,noend]{algorithm2e}

\SetCommentSty{mycommfont}

\usepackage{here}

\usepackage{amsmath,amssymb,amsfonts,amsbsy,amsfonts,latexsym}
\usepackage{makecell}
\usepackage{xcolor}
\usepackage{colortbl}

\usepackage{tabularx,colortbl,xcolor}
\usepackage[normalem]{ulem}
\useunder{\uline}{\ul}{}

\usepackage{enumitem}

\usepackage{xparse}

\SetKwInput{KwInput}{Input}
\SetKwInput{KwRequire}{Require}

\usepackage{longtable}

\NewDocumentCommand{\var}{O{s} m O{}}{%
  \ensuremath{#1_{#2}^{#3}}
}
\usepackage{siunitx}

\newcommand{\commentout}[1]{}

\definecolor{light-gray}{gray}{0.80}

\newcommand{\mrowrot}[2]{
\parbox[t]{2mm}{\multirow{#1}{*}{\rotatebox[origin=c]{90}{#2}}}
}

\newcommand\appref{Appendix~\ref}
\newcommand\eref{Eq.~\ref}
\newcommand\fref{Figure~\ref}
\newcommand\tref{Table~\ref}
\newcommand\sref{Section~\ref}

\usepackage{amsthm}

\usepackage{amsthm}

\newtheorem{mytheorem}{Theorem}
\newtheorem{assumption}[mytheorem]{Assumption}
\newtheorem{lemma}[mytheorem]{Lemma}
\newtheorem{remk}[mytheorem]{Remark}

\def\loss{\mathcal{L}}
\def\lossreg{\loss_{reg}}

\def\lambdareg{\lambda_{reg}}
\def\lambdamax{\lambda_{max}}
\def\lambdamin{\lambda_{min}}

\def\sigmoid{\text{Sigmoid}\xspace}

\def\bertbase{BERT$_{\text{base}}$\xspace}

\newcommand{\OURS}{LEAP\xspace}
\newcommand{\softmvp}{Soft MvP\xspace}

\newcommand{\squad}{SQuAD\xspace}

\usepackage{chngcntr}
\usepackage{adjustbox}

\usepackage{microtype}
\usepackage{graphicx}
\usepackage{subfigure}
\usepackage{booktabs} 

\usepackage{hyperref}

\usepackage[accepted]{icml2022}

\usepackage{amsmath}
\usepackage{amssymb}
\usepackage{mathtools}
\usepackage{amsthm}
\usepackage{bm} 
\usepackage[capitalize,noabbrev]{cleveref}

\usepackage[textsize=tiny]{todonotes}

\icmltitlerunning{Learnable Pruning}

\begin{document}

\twocolumn[
\icmltitle{\OURS: Learnable Pruning for Transformer-based Models}



\icmlsetsymbol{equal}{*}

\begin{icmlauthorlist}
\icmlauthor{Zhewei Yao}{equal,Microsoft}
\icmlauthor{Xiaoxia Wu}{equal,Microsoft}
\icmlauthor{Linjian Ma}{UIUC}
\icmlauthor{Sheng Shen}{UCB}
\icmlauthor{Kurt Keutzer}{UCB}
\icmlauthor{Michael W. Mahoney}{ICSIUCB}
\icmlauthor{Yuxiong He}{Microsoft}
\end{icmlauthorlist}

\icmlaffiliation{Microsoft}{Microsoft, Bellevue, WA}
\icmlaffiliation{UCB}{University of California, Berkeley}
\icmlaffiliation{UIUC}{University of Illinois Urbana-Champaign}
\icmlaffiliation{ICSIUCB}{International Computer Science Institute and University of California, Berkeley}

\icmlcorrespondingauthor{Zhewei Yao}{zheweiyao@microsoft.com}
\icmlcorrespondingauthor{Xiaoxia Wu}{xiaoxiawu@microsoft.com}


\vskip 0.3in
]



\printAffiliationsAndNotice{\icmlEqualContribution} 

\begin{abstract}
Pruning is an effective method to reduce the memory footprint and computational cost associated with large natural language processing models.
However, current pruning algorithms either only focus on one pruning category, e.g., structured pruning and unstructured, or need extensive hyperparameter tuning in order to get reasonable accuracy performance.
To address these challenges, we propose LEArnable Pruning (\OURS), an effective method to gradually prune the model based on thresholds learned by gradient descent. 
Different than previous learnable pruning methods, which utilize $L_0$ or $L_1$ penalty to indirectly affect the final pruning ratio, \OURS introduces a novel regularization function, that directly interacts with the preset target pruning ratio. 
Moreover, in order to reduce hyperparameter tuning, a novel adaptive regularization coefficient is deployed to control the regularization penalty adaptively. 
With the new regularization term and its associated adaptive regularization coefficient, \OURS is able to be applied for different pruning granularity, including unstructured pruning, structured pruning, and hybrid pruning, with minimal hyperparameter tuning. 
We apply \OURS for BERT models on QQP/MNLI/\squad for different pruning settings. 
Our result shows that for all datasets, pruning granularity, and pruning ratios, \OURS achieves on-par or better results as compared to previous heavily hand-tuned methods. 
\end{abstract}
\section{Introduction}
\label{sec:intro}

Since the development of transformer models~\cite{vaswani2017attention}, the number of parameters for natural language processing (NLP) models has become much larger, e.g.,
BERT$_{large}$ (330M)~\cite{devlin2019bert}, 
Megatron-LM (8.3B)~\cite{shoeybi2019megatron}, 
T5 (11B)~\cite{raffel2019exploring},  
GPT3 (170B)~\cite{brown2020language}, 
and MT-NLG (530B)~\cite{mt_nlg}. 
Although larger models tend to exhibit better generalization ability for downstream tasks, the inference time and associated power consumption become critical bottlenecks for deploying those models on both cloud and edge devices. 

One of the promising approaches for addressing the inference time and power consumption issues of these large models is pruning~\cite{sanh2020movement,michel2019sixteen,wang2020spatten}. 
As the nature of neural networks (NNs), different pruning granularity exists, e.g., structured pruning (head pruning for transformers and block-wise pruning for weight matrices) and unstructured pruning (purely sparse-based pruning). 
Different pruning methods are proposed, but they generally only target one set of pruning granularity. 
As such, when a new scenario comes, e.g., hybrid pruning, a combination of structured pruning and unstructured pruning, it is unclear how to choose the proper method. 

Meanwhile, existing work sometimes sets the same pruning ratio for all layers. 
However, it is challenging to prune the same amount of parameters of all weights of a general NNs to ultra-low density without significant accuracy loss. 
This is because not all the layers of an NN allow the same pruning level.
A possible approach to address this is to use different pruning ratios. A higher density ratio is needed for certain ``sensitive'' layers of the network, and a lower density ratio for ``non-sensitive'' layers. 
However, manually setting such multi-level pruning ratios is infeasible. 
Regularization method, e.g.,~\cite{sanh2020movement}, is proposed to address multi-level pruning ratio issue. However, it introduces two drawbacks: (i) a careful hand-tuned threshold schedule is needed to improves the performance, especially in high sparsity regimes; and (ii) due to the regularization term is not directly applied to the final pruning ratio, the regularization magnitude also needs heavily tuning to get desired density ratio.   

Motivated by these issues, we propose an effective LEArnable Pruning (\OURS) method to gradually prune the weight matrices based on corresponding thresholds that are learned by gradient descent.   
We summarize our contributions below,
\begin{itemize}[noitemsep,topsep=0pt,parsep=0pt,partopsep=0pt,leftmargin=*]
\item 
\OURS sets a group of learnable pruning ratio parameters, which can be learned by the stochastic gradient descent, for the weight matrices,  with a purpose to set a high pruning ratio for insensitive layers and vice versa. 
As the NN prefers a high-density ratio for higher accuracy and low loss, we introduce a novel regularization function that can directly control the preset target pruning ratio. 
As such, \OURS can easily achieve the desired compression ratio unlike those $L_0$ or $L_1$ penalty-based regularization methods, whose target pruning ratio needs careful hyperparameter tuning.

\item 
To ease hyperparameter search, we design an adaptive regularization magnitude $\lambda_{reg}$ to adaptively control the contribution to the final loss from the regularization penalty.
The coefficient $\lambda_{reg}$ is automatically adjusted to be large (small) when the current pruning ratio is far away (close to) the target ratio.

\item We apply \OURS for \bertbase on three datasets, i.e., QQP/MNLI/\squad, under different pruning granularity, including structured, hybrid, and unstructured pruning, with various pruning ratios. 
Our results demonstrate that \OURS can consistently achieve on-par or better performance as compared to previous heavily tuned methods, with minimal hyperparameter tuning.

\item We show that \OURS is less sensitive to the hyperparameters introduced by our learnable pruning thresholds, and demonstrate the advance of our adaptive regularization magnitude over constant magnitude. 
Also, by analyzing the final pruned models, two clear observations can be made for BERT pruning: 
(1) early layers are more sensitive to pruning, which results in a higher density ratio at the end;
and (2) fully connected layers are more insensitive to pruning, which results in a much higher pruning ratio than multi-head attention layers.

\end{itemize}
\section{Related Work}
\label{sec:related_work}
Different approaches have been proposed to compress large pre-trained NLP models. 
These efforts can be generally categorized as follows: 
(i) knowledge distillation~\cite{jiao2019tinybert,tang2019distilling,sanh2019distilbert,sun2019patient}; 
(ii) quantization~\cite{bhandare2019efficient,zafrir2019q8bert,shen2020q,fan2020training,zadeh2020gobo,zhang2020ternarybert,bai2020binarybert,esser2019learned};
(iii) new architecture design~\cite{sun2020mobilebert,iandola2020squeezebert,lan2019albert,kitaev2020reformer,wang2020linformer};
and (iv) pruning.
Pruning can  be  broadly  categorized  into unstructured pruning~\cite{dong2017learning,lee2018snip,xiao2019autoprune,park2020lookahead,han2015deep,sanh2020movement} and structured pruning~\cite{luo2017thinet,he2018amc,yu2018nisp,lin2018accelerating,huang2018data,zhao2019variational,yu2021hessian,michel2019sixteen}.
Here, we briefly discuss the related pruning work in NLP. 

For unstructured pruning, \cite{yu2019playing,chen2020lottery,prasanna2020bert,shen2021s} explore the lottery-ticket hypothesis~\cite{frankle2018lottery} for transformer-based models; 
\cite{zhao2020masking} shows that pruning is an alternative effective way to fine-tune pre-trained language models on downstream tasks; and 
\cite{sanh2020movement} proposes the so-called movement pruning, which considers the changes in weights during fine-tuning for a better pruning strategy, and which achieves significant accuracy improvements in high sparsity regimes. 
However, as an extension of~\cite{narang2017block}, \cite{sanh2020movement} requires non-trivial hyperparameter tuning to achieve better performance as well as desired pruning ratio.

For structured pruning, \cite{fan2019reducing,sajjad2020poor} uses LayerDrop to train the model and observes that small/efficient models can be extracted from the pre-trained model; 
\cite{wang2019structured} uses a low-rank factorization of the weight matrix and adaptively removes rank-1 components during training; and 
\cite{michel2019sixteen} tests head drop for multi-head attention and concludes that a large percentage of attention heads can be removed during inference without significantly affecting the performance. 
More recently, \cite{lagunas2021block} extends~\cite{sanh2020movement} from unstructured pruning to block-wise structured pruning. 
As a continuing work of~\cite{sanh2020movement,narang2017block}, hyperparameter tuning is also critical for~\cite{lagunas2021block}.

Although fruitful pruning algorithms are proposed, most methods generally only work for specific pruning scenarios, e.g., unstructured or structured pruning. 
Also, a lot of algorithms either (i) need a hand-tuned threshold (aka pruning ratio) to achieve good performances; or (ii) require careful regularization magnitude/schedule to control the final pruning ratio and retain the model quality. 
Our \OURS is a general pruning algorithm that achieves on-par or even better performance under similar pruning ratio across various pruning scenarios as compared to previous methods, and \OURS achieves this with very minimal hyperparameter tuning by introducing a new regularization term and a self-adaptive regularization magnitude. 


\section{Methodology}
Regardless of pruning granularity, in order to prune a neural network (NN) there are two approaches: (i) one-time pruning~\cite{yu2021hessian,michel2019sixteen} and (ii) multi-stage pruning~\cite{han2015deep,lagunas2021block}. 
The main difference between the two is that one-time pruning directly prunes the NN to a target ratio within one pruning cycle. 
However, one-time pruning oftentimes requires a pre-trained model on downstream tasks and leads to worse performance as compared to multi-stage pruning. 
For multi-stage pruning, two main categories are used: (i) one needs multiple rounds for pruning and finetuing~\cite{han2015deep}; and (ii) another gradually increases pruning ratio within one run~\cite{sanh2020movement,lagunas2021block}. 
Here, we focus on the latter case, where the pruning ratio gradually increases until it  reaches the preset target.

\subsection{Background and Problems of Existing Pruning Methods}
\label{sec:background}

Assume the NN consists of $n$ weight matrices, $\mathcal{W}=\{W_1, \ldots, W_n\}$. 
To compress $\mathcal{W}$, gradual pruning consists of the following two stages:
\begin{itemize}[noitemsep,topsep=0pt,parsep=0pt,partopsep=0pt,leftmargin=*]
    \item (S1) For each $W_i$, we initialize a corresponding all-one mask $M_i$ and denote $\mathcal{M}=\{{M_1,\ldots,M_n}\}$ as the whole set of masks.
    \item (S2) We train the network with the objective function,
    \begin{equation*}
    \small
            \min_{\mathcal{W}}\mathcal{L}_{\text{pure}}(\mathcal{M}\odot\mathcal{W}),
    \end{equation*}
    where $\mathcal{M}\odot\mathcal{W}$ means $W_i\odot M_i$ for all $i=1, \dots n$, 
    and $\mathcal{L}_{\text{pure}}$ is the standard training objective function of the associated task, e.g., the finite sum problem with cross-entropy loss. 
    As the training proceeds, the mask $M_i$ is gradually updated with more zero, i.e., the cardinality, $|M_i|=s^i_{t}$ at iteration $t$, becomes smaller.
\end{itemize}
Here $s^i_{t}$ in (S2) could be a simple linear decaying function or more generally a polynomial function based on the user's requirement. 
Such method is called \textit{hard/soft-threshold pruning}. 
In~\cite{zhu2017prune,sanh2020movement,lagunas2021block}, $s^i_t$ is set to be the same across all the weight matrices, i.e., $s^i_t:=s_t$ and  they use a cubic sparsity
scheduling for the target sparsity $s_f$ given a total iterations of $t_f$:
\begin{equation}
\small{
   s_t= \begin{cases}
      s_{0} & 0 \leq t < t_0,\\
      s_f + (s_0 - s_f)(1-\frac{t-(t_0+t_c)}{t_f-(t_0+t_c)})^3 &  t_0 \leq t < t_f - t_c,\\
      s_f & t\geq t_c. \label{eq:sparse-schedule}
    \end{cases}}
\end{equation}
Although threshold methods achieve reasonably advanced pruning ratios along with high model qualities, they also exhibit various issues.
Here we dive deep into those problems.

\paragraph{Common issues} Both hard- and soft- threshold pruning introduce three hyperparameters: the initial sparsity value $s_0$, the warmup step $t_0$, and the cool-down steps $t_c$. 
As a common practical issue, more hyperparameters need more tuning efforts,
and the question, how to choose the hyperparameters $v_0,~t_0$ and $t_f$, is by no means resolved. 

\paragraph{Issues of hard-threshold pruning}
It is natural for weight matrices to have different tolerances/sensitivities to pruning, which means that a high pruning ratio needs to be applied for insensitive layers, and vice versa. 
However, for hard-threshold pruning, which sorts the weight in one layer by absolute values and masks the smaller portion (i.e., $s^i_t$) to zero, it uses the same pruning ratio across all layers. 
That oftentimes leads to a sub-optimal solution for hard-threshold pruning.

As such instead of using a single $s_t$ schedule, a more suitable way is to use different $s_t^{i},~i=1,\ldots n$ for each weight matrix $W_i$.
However, this leads the number of tuning hyperparameters to be a linear function as the number of weight matrices, i.e., $3n$. 
For instance, there are $3\times6\times12=216$ hyperparameters for the popular NLP model--\bertbase, a 12-layer encoder-only Transformer of which each layer consists of 6 weight matrices ~\cite{devlin2019bert}.
Extensively searching for these many hyperparameters over a large space is impractical. 

Except for the single threshold issue, the hard-threshold method is hard to extend to different pruning scenarios, e.g., block-pruning, head pruning for attention heads, and filter pruning for fully connected layers. 
The reason is that the importance of those structured patterns cannot be simply determined by their sum of absolute values or other norms such as the Euclidean norm. 

\paragraph{Issues of soft-threshold pruning}
One way to resolve the above issues is through soft-threshold methods. 
Instead of using the magnitude (aka absolute value) of the weight matrix to generate the mask, soft-threshold methods introduce a regularization (penalty) function $\mathcal{L}_{reg}(\mathcal{S})$  to control the sparsity of the weight parameters (for instance,  $L_p$-norm, $\mathcal{L}_{reg}=\|\cdot\|_p$, with $p=0$ or $p=1$). 
Here, $\mathcal{S}:=\{S_i\}_{i=1}^n$ and each $S_i$ refers to the associated importance score matrix of $W_i$, which is learnable during training. 
Particularly, (i) this $S_i$ can be adopted to different pruning granularity, e.g., structured and unstructured pruning, and (ii) the final pruning ratio of each weight matrix can be varied thanks to the learnable nature of $S_i$.

For soft-threshold pruning, the mask, $M_i$, is generated by the learnable importance score $S_i$ and $s_t$\footnote{When all $s^i_t$ are the same, we drop the superscript for simplicity.} using the comparison function, $M_i = f(S_i) > s_t$.\footnote{Here both the function $f(\cdot)$ and the comparison are element wise and the comparison returns either 1 or 0.} 
Where $f(\cdot)$ is any function that maps real values to $[0,~1]$.
As $f(S_i)$ will prefer larger values as smaller loss will be introduced to training procedure, a regularization term is added to the training objective,
\begin{align}
    \label{eq:regu-loss}
    \mathcal{L}_{\text{obj}}(\mathcal{M}\odot\mathcal{W})=\mathcal{L}_{\text{pure}}(\mathcal{M}\odot\mathcal{W})+\lambdareg \mathcal{L}_{reg}(f(\mathcal{S}))
\end{align}
The coefficient $\lambdareg$ is used to adjust the magnitude of the penalty (the larger $\lambdareg$, the sparser the $\mathcal{W}$).  
Although soft-threshold pruning methods achieve better performance as compared to hard-threshold pruning methods, it introduces another hyperparameter $\lambdareg$.
More importantly, as the final sparsity is controlled indirectly by the regularization term, it requires sizable laborious experiments to achieve the desired compression ratio. 

\paragraph{Short summary}
For both hard- and soft- threshold pruning, users have to design sparsity scheduling which raises hyperparameters search issues.
Hard-threshold pruning can hardly extend to different pruning granularity, which likely leads to sub-optimal solutions by setting the same pruning ratio for all layers. 
While soft threshold methods could be a possible solution to resolve part of the problems, it introduces another extra hyperparameter, $\lambdareg$, and there are critical concerns on how to obtain the target sparse ratio. 
We address the above challenges in the coming section by designing learnable thresholds with (i) a simple yet effective regularization function that can help the users to achieve their target sparse ratio, and (ii) an adaptive regularization magnitude, $\lambdareg$ to alleviate the hyperparameter tuning.


\subsection{LEAP with A New Regularization}
\label{sec:our_methodology}

In order to address the previously mentioned challenges in~\sref{sec:background}, we propose our LEArnable Pruning (\OURS) with a new regularization. 
We denote the learnable threshold vector $\bm{\sigma}=[\sigma_i, \ldots, \sigma_n]$ and each $\sigma_i$ associates with the tuple $(W_i, M_i, S_i)$. 
With a general importance score $\mathcal{S}$ and learnable threshold vector $\bm{\sigma}$, \OURS can be smoothly  incorporated to Top-k pruning method~\cite{zhu2017prune, sanh2020movement}.\footnote{Our methods can thus be easily applied to magnitude-based pruning methods by setting $\mathcal{S}$ to be identical to $\mathcal{W}$~\cite{han2015learning}.}
 
Recall the Top-K pruning uses the score matrix set $\mathcal{S}$ to compute $\mathcal{M}$, i.e., $ M_i = \text{Top-}K (S_i)$ with $K\in[0,100]$ in a unit of percentage. By sorting the elements of the matrix $S_i$, Top-$K$ set the mask $M_i$ for the top $K\%$ to be 1, and the bottom $(100-K)\%$ to 0. Mathematically, it expresses as
\begin{equation}
\label{eq:topk}
\small
   \text{Top-K}(x) = 
    \begin{cases}
    1~~&\text{$x \in \text{sort}(S_i,K\%)$},\\
    0~~&\text{o.w.},
    \end{cases}
\end{equation}
where $\text{sort}(S_i,K\%)$ contains the Top $K\%$   of the sorted matrix $S_i$. 
Here $K$ is determined by the users, and thus follows various kinds of schedules such as the cubic sparsity scheduling, \eref{eq:sparse-schedule}. 
As described in~\sref{sec:background}, such a schedule usually requires extensive engineering tuning in order to achieve state-of-the-art performance.  
Moreover, in~\cite{zhu2017prune}, the $\text{Top-K}(\cdot)$ threshold is fixed for all weight matrices. 
However, different weight matrices have different tolerances/sensitivities to pruning, meaning that a low pruning ratio needs to be applied for sensitive layers, and vice versa. 
In order to resolve those issues, we propose an algorithm to automatically adjust their thresholds for all weight matrices.
More specifically, we define $K$ as 
\begin{align}
\label{eq:learnable_threshold}
\small
& K(\sigma_i):= 100\cdot k(\sigma_i), \nonumber\\
&  k(\sigma_i) =\sigmoid(\sigma_i/T), \text{~~~for}\quad  i=1,\ldots, n
\end{align}
where the Sigmoid function is used to map $\sigma$ to be in the range of $(0, 1)$. 
$T$ is a temperature value which critically controls the speed of $k$ transitioning from $1$ to $0$ as $\sigma$ decreases.
We remark that $\sigmoid$ could be replaced with any continuous function that maps any positive or negative values to $[0,1]$. Investigating for various such functions could be an interesting future direction.




For a mask $M_i\in \mathbb{R}^{d^i_{\text{in}}\times d^i_{\text{out}}}$, its  density ratio $|M_i|/ (d^i_{\text{in}}\times d^i_{\text{out}})=k(\sigma_i)$ is uniquely determined by $\sigma_i$. 
However, directly applying this for our objective function will tend to make $k(\sigma_i)$ always close to $1$, since the model prefers no pruning to achieve lower training loss. 
Therefore, we introduce a novel regularization term to compensate for this. 
Denote $R(\bm{\sigma})$ the remaining ratio of weight parameter, which is a function of $\bm{\sigma}$ (more details of how to calculate $R(\bm{\sigma})$ are given later).
Suppose that our target pruning ratio is $R_{target}$. 
We propose the following simple yet effective regularization loss, 
\begin{equation}
\label{eq:reg_loss_ours}
\small
\lossreg(\bm{\sigma}) = \begin{cases}
(R(\bm{\sigma})-R_{target})^2 &R(\bm{\sigma})\geq R_{target},\\
0 &\text{else}.
\end{cases}    
\end{equation}
Equipped with \eref{eq:topk}, \ref{eq:learnable_threshold}, and~\ref{eq:reg_loss_ours}, we then rewrite the training objective as
\begin{align}
    \label{eq:final-regu-loss}
    \mathcal{L}_{\text{obj}}(\mathcal{M}_{\bm{\sigma}}\odot\mathcal{W}) =\mathcal{L}_{\text{pure}}(\mathcal{M}_{\bm{\sigma}}\odot\mathcal{W}) +\lambdareg \lossreg(\bm{\sigma})
\end{align}
where the masks $\mathcal{M}_{\bm{\sigma}}$ is written in an abstract manner, meaning that each mask $M_i$ is determined by Top-K (defined in~\eref{eq:topk}).
As the Top-$k$ operator is not a smooth operator, we use the so-called Straight-through Estimator~\cite{bengio2013estimating} to compute the gradient with respect to both $\bm\sigma$ and $\mathcal{S}$. 
That is to say, the gradient through Top-$K$ operator is artificially set to be $1$. 
%

With such a regularization defined in~\eref{eq:final-regu-loss}, there exits ``competition" between $\sigma_i$ in $\mathcal{L}_{\text{pure}}$ and  $\sigma_i$ in $\lossreg$.
Particularly, $\sigma_i$ in $\mathcal{L}_{\text{pure}}$ tends to make $k(\sigma_i)$ close to 1 as the dense model generally gives better accuracy performance, while  $\sigma_i$ in $\lossreg$ makes $k(\sigma_i)$ close to the target ratio $R_{target}$.  
Notably, our regularization method is fundamentally different from those soft-threshold methods by using $L_0$ or $L_1$ regularization. 
While they apply a penalty to the score matrices with indirect control on final sparsity, our method focus on learnable sparsity thresholds $\sigma_i$. 
Thus, we could easily achieve our target compression ratios. 
On the other hand,  one may add $L_0$ or $L_1$ regularization to~\eref{eq:final-regu-loss} as the two are complementary.


\paragraph{Critical term $R(\bm{\sigma})$} \label{para:sigma-calculation}
We now delve into the calculation of $R(\bm{\sigma})$. 
For simplicity, we consider that  all three matrices $M_i$, $W_i$, and $S_i$ follow the same dimensions $d^i_{\text{in}}\times d^i_{\text{out}}$. 
Then 
\begin{equation*}
    R(\bm{\sigma})=N_{\text{remain}}(\bm{\sigma})/N_{\text{total}},
\end{equation*}
where the total number of weight parameters 
$N_{\text{total}}=\sum_{i=1}^n (d^i_{\text{in}}\times d^i_{\text{out}})$,
and the number of remaining parameters 
$N_{\text{remain}}(\bm{\sigma})=\sum_{i=1}^n k(\sigma_i)(d^i_{\text{in}}\times d^i_{\text{out}})$. 

\paragraph{Adaptive regularization coefficient $\lambdareg$} 
Generally, for regularization-based (e.g., $L_1$ or $L_0$ regularization) pruning methods, $\lambdareg$ needs to be carefully tuned~\cite{sanh2020movement}.
To resolve this tuning issue, we propose an adaptive formula to choose the value of $\lambdareg$, 
\begin{equation}
\label{eq:adaptive_lamdba}
\small
    \lambdareg = \max\left\{\lambdamax \frac{\lossreg}{(1-R_{target})^2}, \lambdamin\right\},
\end{equation}
where $\lambdamax$ and $\lambdamin$ are pre-chosen hyperparameters.%
\footnote{Here $\lossreg$ is just used as a scalar value, which does not affect the computation of gradient. 
For instance, in PyTorch/TensorFlow, we need to detach it from the computational graph.} 
We have found that our results are not sensitive to the choice of these hyper-parameters. 
The idea is that when $R(\bm{\sigma})$ is far away from the $R_{target}$, the new coefficient $\lambdareg$ in \eref{eq:adaptive_lamdba} is close to $\lambdamax$ (when $R(\bm{\sigma})=1$, it is indeed $\lambdamax$) so that we can have a strong regularization effect; and when $R$ is close to $R_{target}$, the penalty can be less heavy in \eref{eq:final-regu-loss}. Detailed comparison between constant and our proposed adaptive regularization are referred to~\sref{sec:results}.

\section{Experimental Setup}
\label{sec:experimental_setup}

We apply \OURS with task-specific pruning for \bertbase, a 12-layer encoder-only Transformer model~\cite{devlin2019bert}, with approximately 85M parameters excluding the first embedding layer. 
We focus on three monolingual (English) tasks: question answer (\squad v1.1)~\cite{rajpurkar2016squad};  sentence similarity (QQP)~\cite{iyer2017first}; and natural language inference (MNLI)~\cite{williams2017broad}. 
For \squad, QQP, and MNLI, there are 88K, 364K, 392K training examples respectively.

In order to do a fair comparison with Soft MvP~\cite{lagunas2021block}, for all tasks, we perform logit distillation to boost the performance~\cite{hinton2015distilling}. 
That is, 
\begin{equation}
\label{eq:distillation}
\small
    \loss_{obj} = \alpha\loss_{ds} +   (1-\alpha) \mathcal{L}_{\text{ce}}(\mathcal{M}({\bm{\sigma}})\odot\mathcal{W})+\lambdareg \lossreg(\bm{\sigma})
\end{equation}
Here, $\loss_{ds}$ is the KL-divergence between the predictions of the student and the teacher, $\mathcal{L}_{\text{ce}}$ is the original cross entropy loss function between the student and the true label,
and $\alpha$ is the hyperparameter that balances the cross-entropy loss and the distillation loss. 
We let $\alpha=0.9$ by default for fair comparison (One might be able to improve the results further with more careful hyperparameter tuning and more sophisticated distillation methods). 

\paragraph{Structured/Unstructured/Hybrid pruning}
The basic ingredients of the transformer-based layer consist of multi-headed attention (MHA) and fully connected (FC) sub-layers.  
We denote the sets of weight matrices $\mathcal{W}_{\text{att}}$ for MHA and $\mathcal{W}_{\text{fc}}$ for FC.

Before we give details on the three pruning settings (Structured, Unstructured, and Hybrid), we first explain square $d\times d$ block-wise pruning. 
Consider an output matrix as $W\in\sR^{d_{\text{in}}\times d_{\text{out}}}$, where $d_{\text{in}}=dr$ and $d_{\text{out}}=dc$ (here $r$ and $c$ are integer by design). 
We will define a mask $M$ and a score $S$ with the dimension of  $r\times c$ for the matrix $W$. 
Given a score  $[S]_{i,j}$, if the Top-K operator returns $[M]_{i,j} = 0$, then all $d^2$ elements in the $(i,j)$-th block of $W$ will be set to $0$; 
otherwise, $[M]_{i,j} = 1$ means keeping those elements. 

In our experiments, \textit{structured pruning} refers to applying block-wise pruning to both sets, i.e., $\mathcal{W}_{\text{att}}$ and $\mathcal{W}_{\text{fc}}$. 
In addition, we make the square block size the same in both MHA and FC sub-layers and we choose $d=32$. 
\textit{Unstructured pruning} is using $d=1$ for MHA and FC. 
\textit{Hybrid pruning} means using structured pruning for MHA (setting the block size to be $d=32$) and using unstructured one for FC ($d=1$). 
As such, there are three different sets of experiments and we summarize them in~\tref{table:pruning_methods}.
\begin{table}[tb]
\caption{Summary of different pruning settings. Here the first column shows the abbreviate name we will refer to later, the second column shows the block size used for multi-head attention (MHA), and the third column shows the block size used for fully-connected layers (FC).}
\label{table:pruning_methods}
\centering
\begin{adjustbox}{width=0.6\linewidth}
\begin{tabular}{ @{}lccc }
\toprule
Pruning Settings & MHA  & FC  \\
\midrule
Hybrid ($\text{H}_{32}$)& $32\times 32$ &$1\times 1$ \\
Structure ($S_{32}$) & $32\times 32$ & $32\times 32$  \\
Structure  ($S_{16}$)& $16\times 16$ & $16\times 16$\\
Structure  ($S_{8}$)& $8\times 8$ & $8\times 8$\\
Unstructure  ($S_{1}$) & $1\times1$ &$1\times 1$ \\
\bottomrule
\end{tabular}
\vspace{-1.4cm}
\end{adjustbox}
\end{table}
We test our methods with the scores $\mathcal{S}$ described in movement pruning~\cite{sanh2020movement,lagunas2021block} over three datasets across unstructured, hybrid, and structured pruning setups. 
Moreover, we follow strictly~\cite{lagunas2021block} (referred to as Soft Pruning in later text) on the learning rate including warm-up and decaying schedules as well as the total training epochs to make sure the comparison is fair.
Let \OURS-l and soft MvP-1 denote a double-epoch training setup compared to \OURS and Soft Pruning \cite{sanh2020movement}. 
For more training details, see~\appref{sec:training_details}. 
\section{Results}
\label{sec:results}
In this section, we present our results for unstructured pruning and structured block-wise pruning, and compare with \cite{sanh2020movement,lagunas2021block} and not include other methods for the two reasons: (1)  our training setup are close to them and they are the current stat-of-the-art methods for BERT models. (2) in \cite{sanh2020movement,lagunas2021block}, there already exist extensive comparisons between different methods, including hard-threshold and $L_0$ regularization

\subsection{Unstructured pruning}
\label{sec:unstructure}



\begin{table}[t]
\caption{
Different density ratios for unstructured pruning.
Here Soft MvP is referred to~\cite{lagunas2021block}.
Here \OURS uses exactly training strategies as~\cite{lagunas2021block} and \OURS-l doubles the training epochs. 
For QQP, we report accuracy and F1 socre; for MNLI, we report the accuracy of match and mis-match sets; for \squad, we report exact match and F1 score.
}\centering
\label{tab:unstructured_pruning}
\begin{adjustbox}{width=0.9\linewidth}
\centering
\begin{tabular}{rlccc }
\toprule
& Methods  & Density 1   & Density 2    & Density 3 \\
\midrule
&          & 9$\sim$10\% & 3$\sim$4\%   & 1$\sim$2\% \\
& Soft MvP & 90.2/86.8   & 89.1/85.5    & N/A \\ 
& \OURS    & 90.4/87.1   & 90.3/87.0    & 89.6/86.0 \\
\mrowrot{-3}{QQP}
& \OURS-l  & \textbf{90.9}/\textbf{87.9}   & \textbf{90.6}/\textbf{87.4}    & \textbf{90.4}/\textbf{87.1} \\
\midrule
&          & 12$\sim$13\% & 10$\sim$11\%   & 2$\sim$3\% \\
& Soft MvP & N/A   & 81.2/81.8    & 79.5/80.1 \\ 
& \OURS    & 81.3/81.5   & 80.6/81.0    & 79.0/79.3 \\
\mrowrot{-3}{MNLI}
& \OURS-l  & \textbf{82.2}/\textbf{82.2}   & \textbf{81.7}/\textbf{81.7}    & \textbf{80.3}/\textbf{80.2} \\
\midrule
&          & 9$\sim$10\% & 5$\sim$6\%   & 3$\sim$4\% \\
& Soft MvP & 76.6/84.9   & N/A    & 72.7/82.3 \\ 
& \OURS    & 77.0/85.4   & 74.3/83.3    & 72.9/82.5 \\
\mrowrot{-3}{SQuAD}
& \OURS-l  & \textbf{78.7}/\textbf{86.7}   & \textbf{75.9}/\textbf{84.5}    & \textbf{75.7}/\textbf{84.5} \\
\bottomrule
\end{tabular}
\end{adjustbox}
\end{table}

Unstructured pruning is one of the most effective ways to reduce the memory footprint of NNs with minimal impact on model quality. 
Here, we show the performance of \OURS under different density ratios of \bertbase on QQP/MNLI/\squad. 

As can be seen, compared to \softmvp, \OURS achieves better performances on 4 out of 6 direct comparisons (as \softmvp only provides two pruning ratios per task). 
Particularly, for QQP, \OURS is able to reduce the density ratio to 1$\sim$2\% while achieving similar performance as \softmvp with 3$\sim$4\% density ratio;
for MNLI, although \OURS is slightly worse than \softmvp, the performance gap is within 0.6 for all cases. 
Also, recall that in order to achieve different level of pruning ratios as well as good model quality, \softmvp needs to be carefully tuned $s_t$ in~\eref{eq:sparse-schedule} and $\lambdareg$ in~\eref{eq:regu-loss}. 
However, for \OURS, the tuning is much friendly and  stable (see~\sref{sec:analysis}).

We also list the results of \OURS-l, which utilizes more training epochs to boost the performance, in~\tref{tab:unstructured_pruning}. 
Note that for all 9 scenarios, the performance of \OURS-l is much better than \OURS, particularly for extreme compression. 
For instance, for MNLI (\squad) with 2$\sim$3\% (3$\sim$4\%) density ratio, longer training brings 1.3/0.9 (2.8/2.0) extra performance as compared to \OURS. 
Meanwhile, for all tasks, \OURS-l demonstrates a better performance as compared to \softmvp. 

One hypothesis to explain why longer training can significantly boost the performance of \OURS is that \OURS introduces both more learnable hyperparameters and the adaptive regularization magnitude. 
As such, those extra parameters need more iterations to reach the ``optimal'' values (which is also illustrated in~\sref{sec:analysis}). 


\subsection{Hybrid and structure pruning}
\label{sec:result_of_head_row_pruning}

\begin{table}[t]
\caption{Hybrid and structured pruning comparison between \OURS and \softmvp~\cite{lagunas2021block}.
For QQP, we report accuracy and F1 socre; for MNLI, we report the accuracy of match and mis-match sets; for \squad, we report exact match and F1 score. LEAP-1  (MvP-1)  means the training iterations is twice larger than LEAP (MvP).}
\label{tab:hybrid_structured_pruning}
\begin{adjustbox}{width=\linewidth}
\centering
\begin{tabular}{rlccc }
\toprule
& Methods     & Density 1 & Density 2 & Density 3 \\
\midrule
        &    & 27$\sim$30\% & 21$\sim$25\% & 11$\sim$15\% \\
& soft MvP-1 (H$_{32}$)  & NA/87.6    & NA/87.1    & NA/\textbf{86.8}    \\
\mrowrot{-1.5}{QQP}
& LEAP-1 (H$_{32}$)     & 91.2/\textbf{88.0}  & 91.0/\textbf{87.9}   & 90.7/85.5   \\
\cdashline{2-5}
& LEAP-1 (S$_{32}$)     & 91.0/87.9 & 90.9/87.7 & 90.5/87.3 \\
\midrule
        &   & 27$\sim$30\% & 17$\sim$21\% & 11$\sim$15\% \\
& soft MvP-1 (H$_{32}$)  & \textbf{83.0}/\textbf{83.6}  & \textbf{82.3}/\textbf{82.7} & 81.1/\textbf{81.5}  \\
\mrowrot{-1.5}{MNLI}
& LEAP-1 (H$_{32}$)     & \textbf{83.0}/83.2  & 82.2/82.4 & \textbf{81.5}/\textbf{81.5}  \\
\cdashline{2-5}
& LEAP-1 (S$_{32}$)     & 82.0/82.1  & 81.0/81.0   & 80.0/80.0   \\
\midrule
 &          & 27$\sim$30\% & 21$\sim$25\% & 15$\sim$19\% \\
& soft MvP-1 (H$_{32}$) & \textbf{80.5}/\textbf{88.7}  & \textbf{79.3}/86.9  & \textbf{78.8}/\textbf{86.6} \\
& LEAP-1 (H$_{32}$)     & 80.1/87.6 & \textbf{79.3}/\textbf{87.0}  & 78.2/86.1  \\
\cdashline{2-5}
& soft MvP-1 (S$_{32}$)  & \textbf{77.9}/85.6 & 77.1/85.2 & N/A         \\
\mrowrot{-4}{\squad}
& LEAP-1 (S$_{32}$)     & \textbf{77.9}/\textbf{85.9} & \textbf{77.2}/\textbf{86.4} & N/A         \\
\cdashline{2-5}
& LEAP-1 (S$_{16}$)     & 78.1/86.1 & 77.8/86.1 & N/A        
\\
& LEAP-1 (S$_{8}$)     & 78.7/86.7 & 78.3/86.3 & N/A        
\\
\bottomrule
\end{tabular}
\end{adjustbox}
\vspace{-0.3cm}
\end{table}

We start with hybrid pruning and compare \OURS-1 with \softmvp-1. 
The results are shown in~\tref{tab:hybrid_structured_pruning}. 
Again, as can be seen, for different tasks with various pruning ratio, the overall performance of \OURS-1  is similar to \softmvp-1, which demonstrates the easy adoption feature of \OURS.

We also present structured pruning results in~\tref{tab:hybrid_structured_pruning}. 
The first noticeable finding as expected here is that the accuracy drop of structured pruning is much higher than hybrid mode, especially for a low density ratio. 
Compared to \softmvp-1, \OURS-1  achieves slightly better performance on \squad. For more results,  see~\appref{sec:Results_details}.

We reiterate that  \emph{\tref{tab:unstructured_pruning}  and \tref{tab:hybrid_structured_pruning} are not about  beating the state-of-the-art results but emphasizing that LEAP requires much less hyper-parameter tuning but achieves similar performance as \softmvp that involved a large set of the hyper-parameter sweep. }
For  details about hyper-parameter tuning of \softmvp-1 and LEAP, see \sref{sec:Results_details}.


\section{Analysis}
\label{sec:analysis}
As mentioned, \OURS is a learnable pruning method with a minimal requirement of hyperparameter tuning. 
In order to demonstrate this, 
we analyze \OURS by delving into the key components of LEAP: the initialization of our thresholds $\bm\sigma$, the temperature $T$, and the regularization term $\lambda_{reg}$. 

\begin{figure}[htb]
\centering
\includegraphics[width=0.5\textwidth]{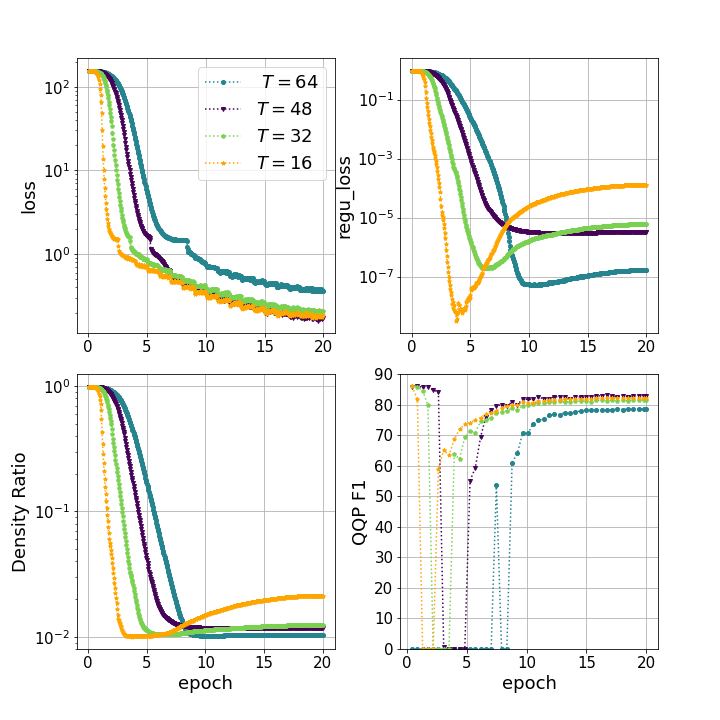}
\vspace{-0.5cm}
\caption{Effect of temperature $T$ for unstructured pruning on QQP. 
The density ratio is set to be 1\%.
}
\label{fig:tempT}
\vspace{-0.5cm}
\end{figure}
\begin{figure}[htb]
\centering
\includegraphics[width=0.5\textwidth]{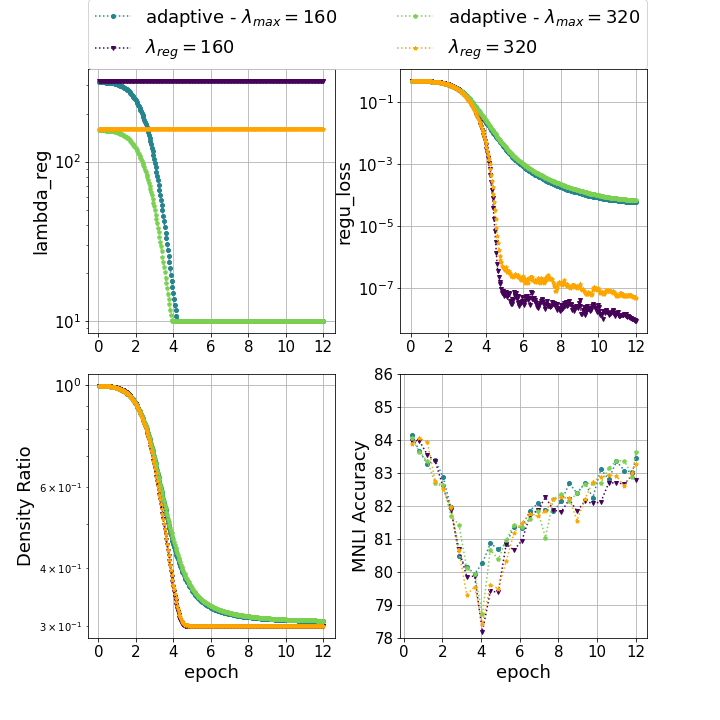}
\vspace{-1.3cm}
\caption{Effect of adaptive regularization $T$ for Hybrid  pruning (H$_{32}$) on MNLI with a target dense ratio of $30\%$. Note that in the plot of density ratio with respect to epochs (left bottom), the purple (blue) and orange (green) curves are overlapped. 
Also in the right top bottom, blue and green curves are overlapped.
}
\label{fig:adaptive}
\vspace{-0.4cm}
\end{figure}
\paragraph{Temperature $T$}  
As $T$ defined in~\eref{eq:learnable_threshold}, it plays a critical role in determining the rate at which the threshold curve $k(\sigma_i)$ falls. 
In addition, $T$ also directly links to the initialization of $\sigma_i$ which is set to be $5T$  for all $i$ such that $\sigmoid(\sigma_i/T)\approx1$. 
This allows the model to have sufficient time to identify the layers which are insensitive for aggressive pruning and vice versa.
To understand how $T$ influences the performances of the Bert model, we conduct an unstructured pruning on the QQP dataset by varying $T\in\{64,48,32,16 \}$ and keeping all other hyperparameters to be the same. 
We plot the objective loss $\mathcal{L}_{\text{obj}}$ (loss), the regularization loss  $\mathcal{L}_{\text{reg}}$ (regu\_loss), the density ratio $R(\bm{\sigma})$, and F1 accuracy, with respect to the iterations in~\fref{fig:tempT}. 

\begin{table}[H]\label{table:robustness}
\caption{Unstructured pruning on \squad with epoch 10 using various values of regularization coefficient $\lambdareg$ in~\eref{eq:final-regu-loss}. 
It shows that our LEAP is not too sensitive to the hyper-parameter choices $T$ and $\lambdareg$.}
 \begin{adjustbox}{width=\linewidth}
\begin{tabular}{c|cc|cc|cc}
\toprule
\multirow{2}{*}{Temperature} & \multicolumn{2}{c|}{$\lambdareg=50$} & \multicolumn{2}{|c|}{$\lambdareg=160$} & \multicolumn{2}{|c}{$\lambdareg=320$} \\ \cline{2-7} 
& acc/f1                  & density             & acc/f1           & density     & acc/f1        & density        \\\midrule
$T=16 $        & 76.58/84.94             & 10.66               & 76.49/85.01      & 10.27       & 76.33/84.85        & 10.2           \\
$T=32 $        & 77.11/85.49             & 11.46               & 76.97/85.47      & 10.39       &76.96/85.36        & 10.23  \\
\bottomrule       
\end{tabular}
\end{adjustbox}
\end{table}
\vspace{-0.3cm}
\begin{figure*}[htb]
\centering
\includegraphics[width=.89\textwidth]{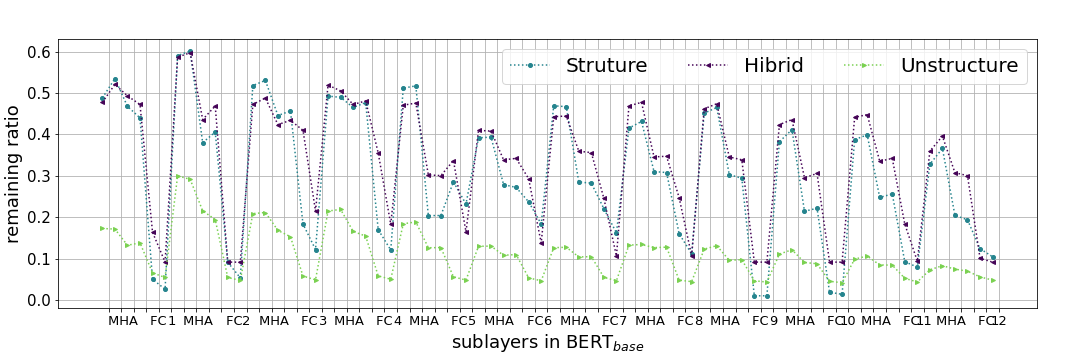}
\caption{The density ratio $k(\sigma_i)$ to  all the weight matrices for structured, hybrid and unstructured pruning on \squad, of which the total density ratios are respectively 20\% , 16\%, and 8\%. 
}
\label{fig:remaining_parameter_ratio}
\vspace{-0.3cm}
\end{figure*}


Among the four curves in~\fref{fig:tempT}, $T=48$ gives the best F1 accuracy while achieving $\sim$1\% density, which clearly demonstrates the significance of $T$ for \OURS. 
Meanwhile, we see that the gaps between the performance for all $T$ except $64$ are close, thus it shows that \OURS is not sensitive to $T$. 
A possible explanation why $T=64$ gives the worse performance is that the density ratio of $T=64$ decays relatively slower compared to rest curves. 
As such, when it is close to the desired pruning regime, the learning rate is relatively small
and so it cannot be able to recover the accuracy. 
On the other hand, it is interesting to note that using the temperature $T=16$ (orange curve), the density ratio increases after around five epochs and keeps increasing to the end\footnote{Please note that the y-axis of the density plot is in logarithmic scale. Even $T=16$ slightly increases the density ratio, it is still very close to 1\%.}, which results in a much better performance even though it experiences the most accuracy drop in the beginning. 
This in some scenes illustrates the ``competition" between $\sigma_i$ in $\mathcal{L}_{\text{pure}}$ and  $\sigma_i$ in $\lossreg$ mentioned in~\sref{sec:our_methodology}: the accuracy increases at epoch $5$ meaning that $\mathcal{L}_{\text{pure}}$ is decreasing effectively and the $\lossreg$ increases (compromises). 
Compared to those manual scheduling thresholds, this increasing phenomena of $\sigma_i$  also shows the advantage of learnable thresholds verifying that the model can figure out automatically when to prune and when not.

\paragraph{Robustness of hyper-parameter tuning $T$ and $\lambdareg$}  
We see in the previous section that given the same $\lambdareg$, various values of the temperature $T$ lead to similar results although tuning is necessary to achieve the best one. 
Here we study how robust the coefficient of $\lambdareg$ in our proposed regularization $\mathcal{L}_{reg}$. 
We prune \bertbase on the SQuAD task with a target ratio $10\%$ with a combination of  $\lambdareg \in \{50, 160, 320\}$ and  $T\in\{16,32\}$, for which the results is in~\tref{table:robustness}. 

For a given $T$, it  indicates that the results are not highly sensitive to different $\lambdareg$s as there is only about 0.1 variation for accuracy. 
It is worth noticing that a smaller  $\lambdareg$ (here $\lambdareg=50$) can indeed affect achieving our target sparse ratio. 
However, the most off pruning ratio is 11.46\%, which is reasonably close to the desired target of 10\%. 

For a given $\lambdareg$, larger $T$ leads both the accuracy and the density ratio higher as expected. 
The reason is that the density ratio function, i.e., Sigmoid$(\sigma_i/T)$, becomes flatter for larger $T$, which leads to a higher density ratio by using the same value of $\sigma$ (Generally, $\sigma$ is negative to achieve $<50\%$ density ratio). 
And higher density ratio results in higher accuracy.

Overall, we can see that \OURS is robust to both $T$ and $\lambdareg$.
\paragraph{The regularization coefficient $\lambdareg$} 
To better understand the effect of adaptive $\lambdareg$ (\eref{eq:adaptive_lamdba}), we set  $\lambdamax\in \{160, 320\}$ and fix $ \lambdamin=10$ (same as~\sref{sec:results}) to prune \bertbase on the MNLI task with a target ratio $30\%$. 
In addition, we also compare this adaptive coefficient with their constant counterparts $\lambdareg \in\{160, 320\}$. 
We plots the $\lambdareg$ (lambda\_reg), the regularization loss $\mathcal{L}_{\text{reg}}$ (regu\_loss), the density ratio $R(\bm{\sigma})$, and accuracy, with respect to the iterations in~\fref{fig:adaptive}. 
First of all, we see that our adaptive coefficient $\lambdareg$ decreases in a quadratic manner and reaching to the $\lambdamin=10$ after 4 epochs, which slows down the pruning activities after $4$ epochs. 
Also, note that the curves of different $\lambdamax$ are actually overlapped with each other, which also indicates that \OURS is not vulnerable to $\lambdareg$.
Meanwhile, as $\lambdareg$ quickly reaches $\lambdamin$, the importance score $\mathcal{S}$ has more time to figure out the pruning parameters for the last small portion.
As such, this slowness can in turn decrease the drop of accuracy and thus eventually recover a much better accuracy than that of the constant regularization.

\paragraph{The effect of learnable pruning for different weight matrices}
As mentioned, the sensitivities of different weight matrices are different.
Therefore, a high pruning ratio should be set for insensitive layers, and a low pruning ratio needs to be used for sensitive layers. 
To demonstrate \OURS can automatically achieve this, we plot the remaining parameters per layer for different pruning granularity on \squad in~\fref{fig:remaining_parameter_ratio}. 
As can be seen, different layers receive different pruning ratios. 
Particularly, 
(i) as compared to MHA layers, FC layers are generally pruned more, which results in a lower density ratio. 
This might indicate that FC layers are less sensitive as compared to MHA layers;
(ii) there is a clear trend that shallow layers (close to inputs) have higher density ratios as compared to deep layers (close to outputs). 
This finding is very intuitive. 
If the pruning ratio is too high for shallow layers, the information loss might be too high, and it is hard for the model to propagate the information to the output layer successfully. 
Therefore, the pruning ratio of shallow layers is smaller.



\section{Conclusions}
\label{sec:conclusions}
In this work, we present \OURS, a learnable pruning framework for transformer-based models. 
To alleviate the hyperparameter tuning effort, \OURS (i) introduces a novel regularization function to achieve desired pruning ratio with learnable pruning ratios for different weight matrices, and (ii) designs an adaptive regularization magnitude coefficient to control the regularization loss adaptively. 
By combining these two techniques, \OURS achieves on-par or even better performance for various pruning scenarios as compared to previous methods. 
Also, we demonstrate that \OURS less sensitive to the newly introduced hyperparameters and show the advance of the proposed adaptive regularization coefficient.
Finally, we also show that there is clear pruning sensitivity associated with the depth and the component of the network.

\bibliography{example_paper}
\bibliographystyle{icml2022}

\clearpage
\appendix
\onecolumn

\section{Training Details}
\label{sec:training_details}

 For all three tasks, the temperature parameter $T$ for $k(\sigma_i)$ is chosen between $\{16, 32, 48, 64\}$ and $\lambda_{\max}$ varies between $\{40, 80, 160, 320\}$ with $\lambda_{\min}=10$. For initialization of $\sigma_i$, we set it to be $5T$. We use a batch size 32, a sequence 128 for QQP/MNLI and  11000/12000 warmup steps (about 1 epoch) for learning rate. As for \squad, we use a batch size 16 and a sequence 384, and we use 5400 warmup steps (about 1 epoch). 
We use a learning rate of 3e-5 (1e-2) for the original weights (for pruning-rated parameters, i.e., $\mathcal{S}$ and $\bm{\sigma}$). We set all the training to be deterministic with the random seed of $17$. All the models are trained using FP32 with PyTorch on a single V100 GPU. Note that these configurations strictly follow the experimental setup in \cite{sanh2020movement, lagunas2021block}; readers could check more details there. For the results in Table 1, the entire epoch using LEAP is 10, 6, and 10, respectively, for QQP, MNLI, and \squad. For the results of LEAP-1 and the results in Table 2, we simply double training epochs correspondingly (i.e., 20, 12, and 20).

\section{Results Details}\label{sec:Results_details}

 \paragraph{Smaller tasks.}Note larger datasets (QQP/MNLI/\squad) to evaluate the performance of the pruning method is very common due to the evaluation robustness.
However, to illustrate the generalization ability of LEAP, we also tested its performance on two smaller datasets STS-B and MPRC, using block pruning with size 32x32. 
The results are shown in Table 1. 
As can be seen, with around 20\% density ratio, LEAP still achieves marginal accuracy degradation compared to baseline.  

\begin{table}[H]
\centering
\caption{ Results  for STS-B (baseline is 88.71) and MRPC (baseline is 87.01\%) with different temperature $T$ and adaptive $\lambda_{max}$ for structure pruning (block size 32x32).
} 
\scalebox{0.8}{
\begin{tabular}{l|lll|l|lll}
\hline 
& \multicolumn{3}{c|}{ Spearman correlation } &  & \multicolumn{3}{c}{Density ratio} \\
STS-B & T=1   & T=2    & T=4   &   & T=1    & T=2   & T=4   \\\hline
$\lambda_{max}=$80    & 85.68 & 85.86 & 85.96 &  & 20.0 & 20.1 & 22.5 \\
$\lambda_{max}=$160   & 85.73 & 85.91  & 86.19 &  & 20.0  & 20.1 & 26.0 \\
$\lambda_{max}=$320   & 85.72 & 86.01  & 86.46 &  & 20.0  & 20.3  & 28.5  \\\hline
& \multicolumn{3}{c|}{Accuracy} &  & \multicolumn{3}{c}{Density ratio} \\
MRPC   & T=1   & T=2    & T=4   &  & T=1    & T=2   & T=4   \\\hline
$\lambda_{max}=$80    & 82.1 & 82.6  & 79.16 &  & 20.0     & 20.0   & 20.3  \\
$\lambda_{max}=$160   & 81.37   & 82.35  & 79.65 &  & 20.0    & 20.0    & 21.3  \\
$\lambda_{max}=$320   & 80.88 & 81.37  & 78.67 &  & 20.0     & 20.0    & 22.7 \\\hline
\end{tabular}}
\end{table}

 \paragraph{Hyper-parameter.} We emphasize again the results of soft mvp is a strong baseline, and our goal is not to purely beat soft mvp from accuracy perspective.
However, their results require extensive hyperparameter tuning (see \href{https://github.com/huggingface/nn_pruning/tree/main/analysis/article/files}{directory}), while ours require to only tune $T$. To show the generalization of the best hyperparameter, we include the results for various $\lambda_{max}$ and $T$ on multiple tasks in \tref{tab:hybrid_structured_pruning}.
Note that when $T$ is fixed, different $\lambda_{max}$ gives similar results over various tasks.
\begin{table}[H]
\centering
\caption{ Results for QQP and MRPC with different temperature $T$ and adaptive $\lambda_{max}$ for structure pruning (block size 32x32).
} 
\scalebox{0.8}{
\begin{tabular}{l|lll|l|lll}
\hline 
& \multicolumn{3}{c|}{ Accuracy} &  & \multicolumn{3}{c}{Density ratio} \\
QQP & T=16        & T=32        & T=48        &  & T=16  & T=32  & T=48    \\\hline
$\lambda_{max}$=160  & 90.68       & 90.87       & 90.7        &  & 20.07 & 20.07 & 20.1  \\
$\lambda_{max}$=320  & 90.79       & 90.78       & 90.6        &  & 20.08 & 20.06 & 20.1   \\\hline
& \multicolumn{3}{c|}{ Accuracy (MNLI/MNLI-MM)} &  & \multicolumn{3}{c}{Density ratio} \\
MNLI   & T=16        & T=32        & T=48        &  & T=16  & T=32  & T=48    \\\hline
$\lambda_{max}$=40   & 80.41/81.06 & 80.79/81.12 & 80.87/81.22 &  & 21.07 & 21.52 & 22.25 \\
$\lambda_{max}$=160  & 80.56/80.98 & 80.88/80.81 & 81.02/81.17 &  & 21.07 & 21.46 & 22.15\\ \hline
\end{tabular}}

\end{table}

\end{document}